\documentclass[]{interact}

\usepackage{epstopdf}
\usepackage[caption=false]{subfig}

\theoremstyle{plain}

\theoremstyle{definition}

\theoremstyle{remark}

\usepackage{subfig}
\usepackage{setspace}
\usepackage{float}
\usepackage{graphicx}
\usepackage{amssymb}
\usepackage{tikz}
\usetikzlibrary{shapes.geometric,arrows}
\tikzstyle{startstop} = [rectangle, rounded corners, minimum width=3cm, minimum height=1cm,text centered, draw=black, fill=red!30]
\tikzstyle{io} = [trapezium, trapezium left angle=70, trapezium right angle=110, minimum width=3cm, minimum height=1cm, text centered, draw=black, fill=blue!30]
\tikzstyle{process} = [rectangle, minimum width=3cm, minimum height=1cm, text centered, draw=black, fill=orange!30]
\tikzstyle{decision} = [diamond, minimum width=3cm, minimum height=1cm, text centered, draw=black, fill=green!30]
\tikzstyle{arrow} = [thick,->,>=stealth]

\begin{document}

\articletype{}

\title{Theoretical Analysis of an XGBoost Framework for Product Cannibalization}

\author{
\name{Gautham Bekal and Mohammad Bari}
\affil{T-Mobile, Two Newport, 3625 132nd Ave SE, Bellevue, Washington, 98006, USA}
}

\maketitle
\begin{abstract}
This paper is an extension of our work~\cite{ThreeStageXGB} where we presented a three-stage XGBoost algorithm for forecasting sales under product cannibalization scenario. Previously we developed the model based on our intuition and provided empirical evidence on its performance. In this study we would briefly go over the algorithm and then provide  mathematical reasoning behind its working.

\end{abstract}

\section{Introduction}

The three-stage XGBoost model discussed in previous study had superior performance compared to squared error based XGBoost model when forecasting sales under product cannibalization scenario.
The experiments also showed that the three-stage XGBoost framework was able to capture long-term forecasting dynamics very well. However, in our previous study we lacked mathematical rigor for the algorithm. In this paper we provide theoretical reasons for choosing three-stages in our framework.

\section{Background}
From our previous study we obtained important equations for three-stage XGBoost algorithm which have been briefly described below. We have made some minor changes in the notation here to simplify the understanding, such as replacing ${Y_i}$ by ${Y_{stageNumber_i}}$ to make it more explicit about predictions made by specific XGBoost model. The train set consists of \emph{m} samples and test set is indexed as \emph{m+1} to \emph{n} samples.
The objective is to make predictions of sales for existing products in future weeks i.e. test data when new products are being launched. The primary assumption is that a new product launch will either decrease or increase sale of individual products, however sum of sales of all products for a given week \emph{i} will itself be constant and is obtained as per below equation.
\begin{equation}\label{equation_one}
\begin{split}
Y_{categoryActual_i}=Y_{a_i} + Y_{a_{i+1}} +Y_{a_{i+2}}+.....+Y_{a_{i+count_i}} \;... for \; historical data \\
S_i = domainKnowledge  \;...  \; for \;future \;weeks
\end{split}
\end{equation}
Where, $Y_{a_i}$ is the actual sales of a product in the specific week \emph{i} and $count_i$ is the number of devices in the given week in a given category.
In historical data when actual sales of products is available, total sale in week \emph{i} is obtained by summing over sales of products in that week.
Now, for future weeks where individual product sales is unavailable, we make a fundamental assumption that, the aggregate sale of the category for a given week \emph{i} is easily obtainable by domain knowledge $S_i$.

The objective function of XGBoost1 was given by,
\begin{equation}\label{equation_two}
\emph{$l_1$}=\frac{\sum_{i=1}^{m}(Y_{a_i} - Y_{StageOnePrediction_i})^2}{m}.       
\end{equation}

Where $m$ is the number of training samples with actual sales. $Y_{a_i}$ is the weekly sale of an individual product. $Y_{StageOnePrediction_i}$ is the predicted weekly sale by the XGBoost1 for an individual product. 

XGBoost2 was trained on the below objective function,

 \begin{equation}\label{equation_three}
\begin{split} 
\emph{$l_2$}=\frac{\sum_{i=1}^{m}(Y_{a_i} - Y_{StageTwoPrediction_i})^2 +\sum_{i=m+1}^{n}(Y_{StageOnePrediction_i} - Y_{StageTwoPrediction_i})^2 } {n} \\ + \frac{\sum_{i=1}^{n}(Y_{categoryActual_i}  -Y_{categoryPrediction_i} )^2}{count_i * n}
\end{split}
 \end{equation}

In equation \ref{equation_three}, \emph{$Y_{categoryActual_i}$} is given by,
 \begin{equation}\label{equation_four}
 \emph{$Y_{categoryActual_i}$}=\sum_{i=1}^{count_i}Y_{a_i\;  \; for \;i=1, 2, 3, ...., m.} 
  \end{equation}
  
Where, $Y_{categoryActual_i}$ is the aggregate sum of sales for a given category for week \emph{i} and is obtained as per equation \ref{equation_one}. \emph{$Y_{categoryPrediction_i}$} in equation \ref{equation_three} is given by,

 \begin{equation}\label{equation_five}
\emph{$Y_{categoryPrediction_i}$}=\sum_{i=1}^{count_i}Y_{StageTwoPrediction_i} 
   \end{equation}
 where $Y_{StageTwoPrediction_i}$ is the aggregate sum of predictions made by XGBoost2. Here, $count_i$ is the  number of products on sale in the given week \emph{i}. Thus XGBoost2 was trained on train data indexed 1 to \emph{m} as well as the predictions made by XGBoost1 for test data indexed \emph{m}+1 to \emph{n}.
 
XGBoost3 was trained on the below objective function.
 \begin{equation}\label{equation_six}
 \begin{split} 
 \emph{$l_3$}= \frac{\sum_{i=1}^{n}(Y_{categoryActual_i}  -Y_{categoryPrediction_i} )^2}{count_i * n} + \\ \frac{\sum_{i=1}^{n}(Y_{StageThreePrediction_i} - Y_{predRatio_i}*Y_{categoryActual_i})}{n}
 \end{split}
 \end{equation}
 Here, $Y_{categoryActual_i}$ is obtained similar to equation \ref{equation_four}. $Y_{categoryPrediction_i}$ is the the aggregate sum of predictions produced by XGBoost3 for week \emph{i}, and is similar to \ref{equation_five}. $Y_{predRatio_i}$  is obtained from the predictions of XGBoost1. It is given by,
\begin{equation}\label{equation_seven}
Y_{predRatio_i}=\frac{Y_{StageOnePrediction_i} } {Y_{StageOneCategoryPrediction_i} } =  \frac{Y_{StageOnePrediction_i} } {\sum_{i=1}^{count_i}Y_{StageOnePrediction_i} }
 \end{equation}.

\section{Mathematical Justification}
The important assumption we make is that aggregate sum of product sales is available to us on a weekly basis even for future data via domain knowledge as defined in equation \ref{equation_one}. Aggregate sum of prediction by XGBoost1 model on test data for a given week can be written as,
 \begin{equation}\label{equation_eight}
\emph{$S_{predict_i}$}=f_1(X_{(i+1)}) + f_1(X_{(i+2)}) +...+f_1(X_{(i+count)})
 \end{equation}.
 Where, $S_{predict_i}$ is the sum of predictions for a given week by the XGBoost1 model. $f_1$ is the trained XGBoost1 model. $X_{(m+1)_i},X_{(m+2)_i},...X_{(m+count)_i}$ are the input features of $product_1$, $product_2$...$product_{count}$ on future week \emph{i}. Here, ${count_i}$ is the number of products in the category that contribute to sales of a specific week \emph{i}. However, the predictions made by XGBoost1 has no information about the domain knowledge. Thus ${S_{predict_i}}$ is an unconstrained categorical prediction. Hence,
  \begin{equation}\label{equation_nine}
 {S_{predict_i}} \ne  {S_i}
 \end{equation}
 Our objective is to minimize the difference between ${S_{predict_i}}$ and ${S_i}$, which can be written as,
   \begin{equation}\label{equation_ten}
 min \; ({S_{predict_i}} -  {S_i})
 \end{equation} 
   \begin{equation}\label{equation_eleven}
 i.e  \; ({S_{predict_i}} -  {S_i}) = 0
 \end{equation}
 Squaring on both sides we get,
   \begin{equation}\label{equation_tweleve}
  ({S_{predict_i}} -  {S_i})^2 =0
 \end{equation}
 From equation \ref{equation_eight} and \ref{equation_tweleve} we can write,
    \begin{equation}\label{equation_thirteen}
 ( f_1(X_{(i+1)}) + f_1(X_{(i+2)}) +...+f_1(X_{(i+count_i)}) -  {S_i})^2=0
 \end{equation}
Equation \ref{equation_thirteen} is basically the second term in \ref{equation_three} written in a different form without $count_i$ for normalization. Also, the denominator term \emph{n} is missing since we have not written in the form to take average over the dataset.
 Now, XGBoost2 is trained on \emph{n} samples of dataset where, \emph{m}+1 to \emph{n} samples were predicted by XGBoost1.
If we optimize XGBoost2 on equation \ref{equation_thirteen} only without considering the 1st term from equation \ref{equation_three} then it leads to trivial solution as below,
\begin{equation}\label{equation_fourteen}
  {f_2(X_{(i+1)})} = {f_2(X_{(i+2)})} =....={f_2(X_{(i+count_i)})} = \frac{S_i}{count_i}=constant
 \end{equation}
 Equation \ref{equation_fourteen} would mean that XGBoost2 would have predicted \emph{constant} value for all products in a given week based on aggregate sales $S_i$, independent of input features \emph{X}. We obviously do not want this, since we want our model to learn both input features \emph{X} and simultaneously adhere to the constraint $S_i$ respectively.
 Now we know that XGBoost1 was trained on input feature \emph{X} as per equation \ref{equation_two} and made predictions on test set from \emph{m}+1 to \emph{n} samples as per the algorithm discussed in our previous paper.
 Thus, we can transfer this information from XGBoost1 to train our XGBoost2 model, while also making sure it adheres to the constraint $S_i$.
 Thus from equation, \ref{equation_three} and \ref{equation_thirteen} we can write as,
 \begin{equation}\label{equation_fifteen}
\sum_{i=1}^{m}({Y_{a_i}-f_2(X_i)})^2 + \sum_{i=m+1}^{n}({Y_{StageOnePrediction_i}-f_2(X_i)})^2 
 \end{equation}
The entire equation \ref{equation_fifteen} is the 1st term of equation \ref{equation_three} that has been expanded between data points 1 to \emph{m} and \emph{m}+1 to \emph{n}.
As can be seen from \ref{equation_fourteen} for the historical data between 1 to \emph{m} data-points we already know the actuals {$Y_{a_i}$}. And for future data points \emph{m}+1 to \emph{n} data-points we use predictions made by XGBoost1 model, i.e. {$f_1(X)$}. Thus the two terms in \ref{equation_fifteen} constitute the squared error term in equation \ref{equation_three} respectively.
Combining equation \ref{equation_fourteen} and \ref{equation_fifteen} we get \ref{equation_three}, which is the objective function for XGBoost2 model.
If the dataset is sufficiently large then the XGBoost1 model trained on squared error will have less bias in prediction. However, in our experiments we observe that the XGBoost1 model tends to consistently under forecast or overforecast. 
This would mean that,
\begin{equation}\label{equation_sixteen}
 \; Y_{a_i}>>Y_{StageOnePrediction_i} or Y_{a_i}<<Y_{StageOnePrediction_i}
 \end{equation}
 The implication of equation \ref{equation_fifteen} and \ref{equation_sixteen} on equation \ref{equation_three} is that,
 \begin{equation}\label{equation_seventeen}
{\sum_{i=m+1}^{n}(Y_{StageOnePrediction_i} - Y_{stageTwoPrediction_i})^2} >> {\sum_{i=m+1}^{n}(Y_{categoryActual_i}  -Y_{categoryPrediction_i} )^2}
 \end{equation}
 When, XGBoost2 model is trained on \ref{equation_three} with the bias as seen in \ref{equation_sixteen}, it would cause the model to neglect the 2nd term in equation \ref{equation_six}. This is because 2nd term in equation \ref{equation_six} is too small compared to 1st term. The implication is that model would ignore the sum constraint term we introduced to handle cannibalization, and would simply just learn from input matrix \emph{X}. 
 Hence, to overcome the above issues, we need XGBoost3 model that fine-tunes the predictions made by XGBoost2 model. The XGBoost3 model is trained on the entire dataset of \emph{n} samples similar to XGBoost2 model.
 The XGBoost3 model takes prediction made by XGBoost2 model as a part of input matrix \emph{X} in back period filling phase as discussed in previous paper. However, XGBoost2 predictions are not appended to the target variable {$Y_a$} when training for XGBoost3 model. 
 The XGBoost3 model is trained on objective function defined in equation \ref{equation_six}. The first term in \ref{equation_six} is similar to equation \ref{equation_three} which is the sum constraint term. The 2nd term in \ref{equation_six} is meant for fine tuning purpose.
 This is as per equation \ref{equation_seven} which gives amount of contribution of sales prediction by each product with respect to overall category prediction $Y_{categoryPrediction_i}$ made by XGBoost1. Also from equation \ref{equation_one}, $S_i$ is obtained from domain knowledge and is assumed to be accurate. Now, even though XGBoost1 model might have under or over-forecasted as per equation \ref{equation_sixteen} the $Y_{predRatio_i}$ in equation 
 \ref{equation_seven} is itself quite accurate. Hence, from equation \ref{equation_six} and \ref{equation_seven} we can write the target variable for XGBoost3 as,
  \begin{equation}\label{equation_eighteen}
target = \frac{Y_{StageOnePrediction_i}*{Y_{categoryActual_i}}} {Y_{StageOneCategoryPrediction_i}}
 \end{equation}
Hence, the 1st part of equation \ref{equation_six} is used to make sure sum of sales adheres to the constraint similar to equation \ref{equation_three}. The 2nd part of equation \ref{equation_six} helps to make sure sale forecast of individual products will not over or under forecasted by XGBoost3 and is fine-tuned according to equation \ref{equation_seven}.

\section{Conclusion}
In this work, we have provided mathematical justification for the three-stage XGBoost framework which had been used in our previous study to forecast sales under product cannibalization scenario. Here, the rigorous mathematical reasoning gives us more confidence in our algorithm for forecasting purposes. In the upcoming work we would carry out more experiments especially on forecasting sales of products with low volume.

\nocite{*}
\bibliographystyle{plain}
\bibliography{interactapasample.bib}
\end{document}